\documentclass{sig-alternate}

\usepackage{url}
\usepackage{booktabs}

\begin{document}
%
% --- Author Metadata here ---
\conferenceinfo{KDD '15}{August 11--14, 2015, Sydney, NSW, Australia}
\CopyrightYear{2015} % Allows default copyright year (20XX) to be over-ridden - IF NEED BE.
\crdata{978-1-4503-3664-2/15/08}  % Allows default copyright data (0-89791-88-6/97/05) to be over-ridden - IF NEED BE.
% --- End of Author Metadata ---

\title{Bayesian Poisson Tensor Factorization for Inferring
  Multilateral Relations from Sparse Dyadic Event Counts}

\numberofauthors{2} %  in this sample file, there are a *total*
\author{
% 1st. author
\alignauthor
Aaron Schein\\
       \affaddr{College of Information and Computer Sciences}\\
       \affaddr{University of Massachusetts Amherst}\\
%       \affaddr{Amherst, MA 01003}\\
       \email{aschein@cs.umass.edu}
% 2nd. author
\alignauthor
John Paisley\\
       \affaddr{Department of Electrical Engineering}\\
       \affaddr{Columbia University}\\
%       \affaddr{New York, NY 10027}\\
       \email{jpaisley@columbia.edu}
\and  % use '\and' if you need 'another row' of author names
% 3rd. author
\alignauthor
David M. Blei\\
       \affaddr{Department of Computer Science \& \\Department of Statistics}\\
       \affaddr{Columbia University}\\
%       \affaddr{New York, NY 10027}\\
       \email{david.blei@columbia.edu}
% 4th. author
\alignauthor Hanna Wallach\\
       \affaddr{Microsoft Research \&}\\
       \affaddr{College of Information and Computer Sciences}\\
       \affaddr{University of Massachusetts Amherst}\\       
%       \affaddr{New York, NY 10011}\\
       \email{wallach@microsoft.com}
}

\maketitle
\begin{abstract}
We present a Bayesian tensor factorization model for inferring latent
group structures from dynamic pairwise interaction patterns. For
decades, political scientists have collected and analyzed records of
the form ``country $i$ took action $a$ toward country $j$ at time
$t$''---known as \emph{dyadic events}---in order to form and test
theories of international relations. We represent these event data as
a tensor of counts and develop Bayesian Poisson tensor factorization
to infer a low-dimensional, interpretable representation of their
salient patterns. We demonstrate that our model's predictive
performance is better than that of standard non-negative tensor
factorization methods. We also provide a comparison of our variational
updates to their maximum likelihood counterparts. In doing so, we
identify a better way to form point estimates of the latent factors
than that typically used in Bayesian Poisson matrix
factorization. Finally, we showcase our model as an exploratory
analysis tool for political scientists. We show that the inferred
latent factor matrices capture interpretable multilateral relations
that both conform to and inform our knowledge of international
affairs.
\end{abstract}

\category{I.5.1}{Pattern Recognition}{Models}
\category{J.4}{Computer Applications}{Social and Behavioral Sciences}

\terms{Algorithms, Experimentation}

\keywords{Poisson tensor factorization, {B}ayesian inference, dyadic
  data, international relations}

\section{Introduction}

Social processes are characterized by pairwise connections between
actors, such as people, organizations, corporations, and countries. In
some social processes, actors declare their connections and
researchers can directly study them---e.g., friendships on Facebook or
co-authorships in academia. In other processes, however, these
connections are not explicitly declared.  Rather, they are evidenced
over time via dynamic interaction patterns.  Inferring social
processes from such implicit data is a challenging and important task.

This task is especially motivated in international relations.  For
decades, scholars have collected and analyzed records of pairwise
interactions between countries of the form ``country $i$ took action
$a$ toward country $j$ at time $t$,'' known as \emph{dyadic events}.
These data sets, e.g.,~\cite{singer94correlates}, which are
traditionally small and well-curated, help them form and test theories
of international relations, which often concern the multilateral
behavior of groups of countries. Recently, there has been new interest
in studying less structured, larger scale sources of pairwise
interaction data. Researchers have created several large data sets,
e.g.,~\cite{leetaru13gdelt}, by automatically extracting and encoding
dyadic events from Internet news archives.

These modern data sets differ substantially from their smaller
counterparts, which previously dominated the field. Rather than
documenting high-level, aggregate behaviors, such as formal wars and
military alliances, they document micro-level behaviors at a
day-to-day granularity. Although this new view of the world
potentially paints a more accurate and nuanced picture of
international relations, these data are too noisy and disaggregated to
analyze effectively using traditional techniques. We need new methods
to uncover the latent multilateral relations that underlie these events.

\begin{figure*}[ht]
\label{fig:example}
  \begin{center}
    \includegraphics[width=0.49\linewidth]{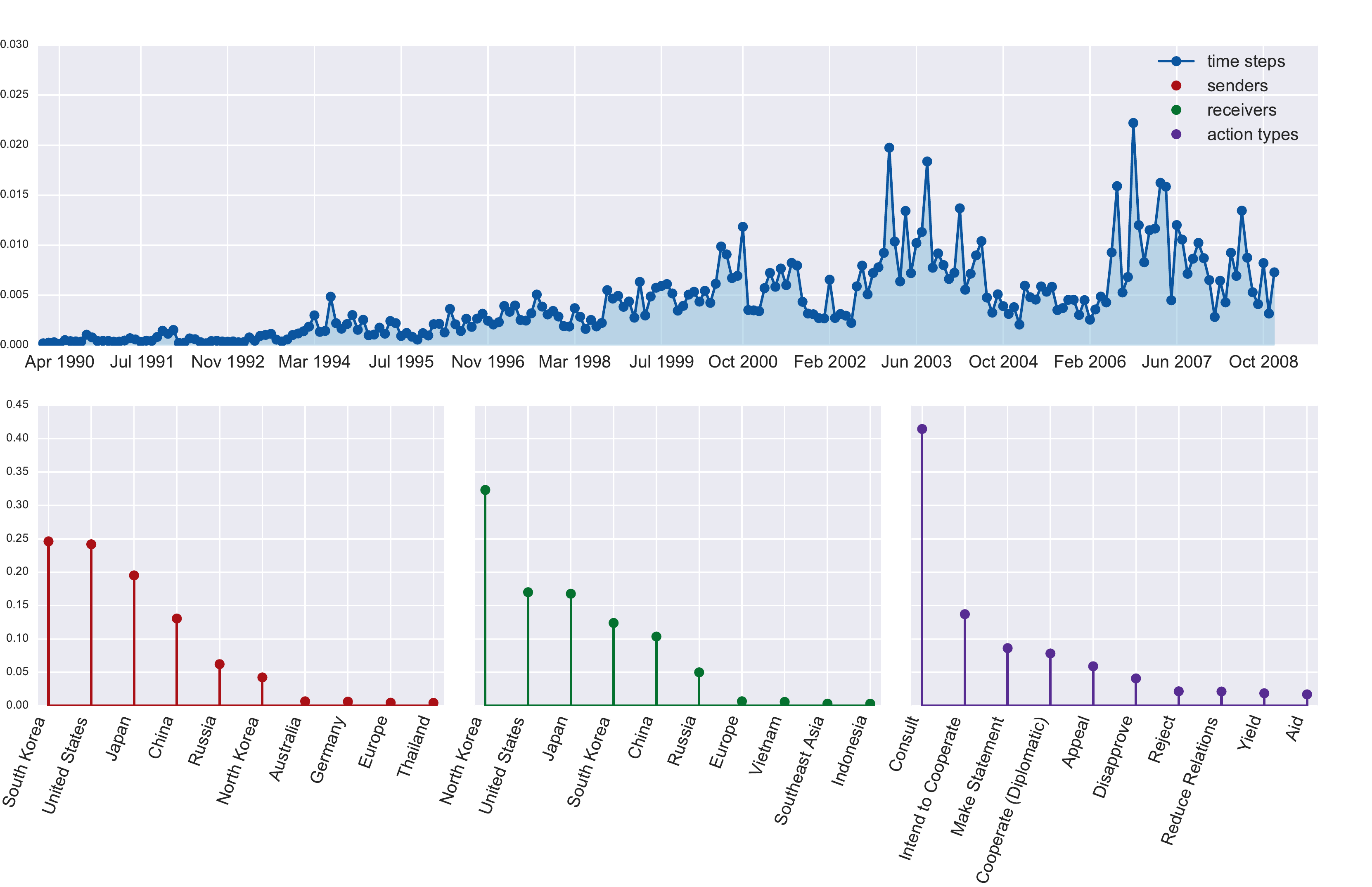}
    \includegraphics[width=0.49\linewidth]{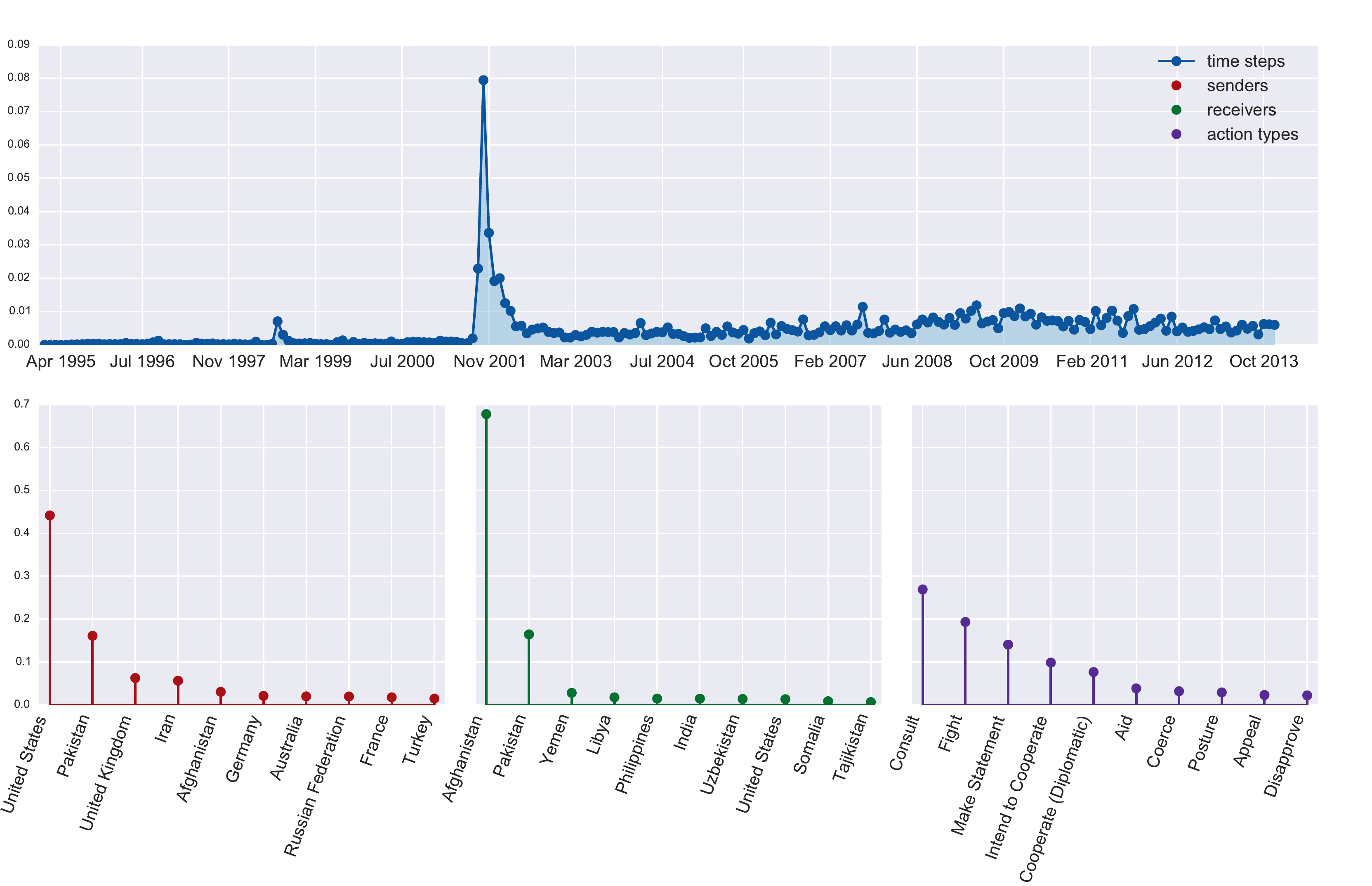}
  \end{center}
  \vspace{-2em}
  \caption{\scriptsize Our model infers latent components that
    correspond to multilateral relations. Each component consists of
    four factor vectors summarizing sender, receiver, action-type, and
    time-step activity, respectively. Here, we visualize two inferred
    components. For each component, we plotted the top ten sender,
    receiver, and action-type factors sorted in decreasing order. We
    also plotted the entire vector of time-step factors in
    chronological order. We found that the interpretation of each
    component was either immediately clear from our existing knowledge
    or easy to discover via a web search. \emph{Left}: A component
    inferred from GDELT data spanning 1990 through 2007 (with monthly
    time steps) that corresponds to events surrounding the Six Party
    Talks---a series of diplomatic talks that took place from 2003
    through 2009 between South Korea, North Korea, the US, China,
    Japan, and Russia, aimed at resolving international concerns over
    North Korea's nuclear weapons
    program~\cite{wikipedia15six-party}. The top senders and receivers
    are the six parties, while the top action types are \emph{Consult}
    and \emph{Intend to Cooperate}. The time-step factors show
    increased activity beginning in 2003. \emph{Right}: A component
    inferred from ICEWS data spanning 1995 through 2012 (with monthly
    time steps) that corresponds to events surrounding the US-led War
    on Terror following the September 11, 2001 attacks. The largest
    time-step factor is that of October 2001---the month during which
    the invasion of Afghanistan occurred. There is also a blip
    in August 1998, when the Clinton administration ordered
    missile attacks on terrorist bases in
    Afghanistan~\cite{wikipedia15cruise}.}
  \vspace{-1em}  
\end{figure*}

In this paper, we introduce Bayesian Poisson tensor factorization
(BPTF) for inferring latent multilateral relations from observed
dyadic events. We present a scalable variational inference algorithm
and demonstrate our method, via both predictive and exploratory
analyses, on large-scale international relations
data. Figure 1 illustrates our approach; our model
infers both ongoing multilateral relations, such as the Six-Party
Talks from 2003 through 2009 (left), as well as relations precipitated
by temporally localized anomalous activity, such as the September 11,
2001 attacks (right).

\vspace{1em}

\noindent \textbf{Technical summary: \,} A data set of dyadic events
can be represented as a four-way tensor by aggregating (i.e.,
counting) events within discrete time steps. Each element of the
tensor is a count of the number of actions of type $a$ taken by
country $i$ toward country $j$ at time $t$. Our model decomposes such
a tensor into a set of latent factor matrices that provide a
low-dimensional representation of the salient patterns in the
counts---in this case, latent multilateral relations.

Tensors derived from dyadic event data are often very sparse since
most countries rarely interact with one another. Additionally, the
non-zero counts, for countries that do interact, are highly
dispersed---i.e., their mean is greatly exceeded by their
variance. Traditional tensor factorization methods, involving maximum
likelihood estimation, are unstable when fit to sparse count
tensors~\cite{chi12tensors}. Bayesian Poisson tensor factorization
(section~\ref{sec:BPTF}) builds on previous work on Bayesian Poisson
matrix factorization with Gamma priors~\cite{cemgil09bayesian,
  liang15codebook-based, gopalan15scalable, zhou15negative,
  paisley14bayesian, gopalan12scalable} to avoid these instabilities.
We validate our model by comparing its out-of-sample predictive
performance to non-Bayesian tensor factorization methods
(section~\ref{sec:eval}); BPTF significantly outperforms other models
when decomposing sparse, highly dispersed count data.

We present an efficient variational inference algorithm to fit BPTF to
data (section~\ref{sec:variational}) and outline the relationship
between our algorithm and the traditional maximum likelihood approach
(section~\ref{sec:discussion}). This relationship explains why BPTF
outperforms other methods without any sacrifice to efficiency. It also
suggests that when constructing point estimates of the latent factors
from the variational distribution, researchers should use the
geometric expectation instead of the arithmetic expectation commonly
used in Bayesian Poisson matrix factorization. We show that using the
geometric expectation increases the sparsity of the inferred factors
and improves predictive performance. We therefore recommend its use in
any subsequent work involving variational inference for Bayesian
Poisson matrix or tensor factorization.

Finally, we showcase Bayesian Poisson tensor factorization as an
exploratory analysis tool for political scientists
(section~\ref{sec:exploratory}). We demonstrate that the inferred
latent factor matrices capture interpretable multilateral relations
that conform to and inform our knowledge of international affairs.

\section{Dyadic Events}
\label{sec:dyadic}

%Political scientists have traditionally used \emph{dyadic
%  events}---i.e., records of pairwise interactions between countries
%of the form, ``country $i$ took action $a$ toward country $j$ at time
%$t$''---to test theories in international relations, often by
%regressing these events against particular country covariates, such as
%military spending or commitment to democratic values.

Over the past few years, researchers have created large data sets of
dyadic events by automatically extracting them from Internet news
archives. The largest of these data sets is the Global Database of
Events, Location, and Tone (GDELT), introduced in 2013, which contains
over a quarter of a billion events from 1979 to the present, and is
updated with new events daily~\cite{leetaru13gdelt}. In parallel,
government agencies (e.g., DARPA) and their contractors have also
started to collect and analyze dyadic events, in order to forecast
political instability and to develop early-warning
systems~\cite{o'brien10crisis}; Lockheed Martin publicly released the
Integrated Crisis Early Warning System (ICEWS) database in early
2015. Ward et al. provide a comparison of GDELT and
ICEWS~\cite{ward13comparing}.

GDELT and ICEWS use the CAMEO coding scheme~\cite{gerner02conflict}. A
CAMEO-coded dyadic event consists of four pieces of information: a
sender, a receiver, an action type, and a timestamp. An 
example of such an event (top) and a sentence from which it
could have been extracted (bottom) is
$$\langle \textrm{Turkey, Syria, \emph{Fight}, } 12/25/2014 \rangle$$
\begin{center}
Dec. 25, 2014: ``Turkish jets bombed targets in Syria.''
\end{center}
CAMEO assumes that senders and receivers belong to a single set of
actors, coded for their country of origin and sector (e.g., government
or civilian) as well as other information (such as religion or
ethnicity). CAMEO also assumes a hierarchy of action types, with the
top level consisting of twenty basic action classes. These classes are
loosely ranked based on sentiment from \emph{Make Public Statement} to
\emph{Use Unconventional Mass Violence}. Each action class is
subdivided into more specific actions; for example, \emph{Make Public
  Statement} contains \emph{Make Empathetic Comment}. When studying
international relations using CAMEO-coded data, researchers commonly
consider only the countries of origin as actors and only the twenty
basic action classes as action types. In ICEWS, there are 249 unique
country-of-origin actors (which include non-universally recognized
countries, such as Taiwan and Palestine); in GDELT, there are 223.

\vspace{0.5em}

\noindent \textbf{Dyadic events as tensors:\,} A data set of dyadic
events can be aggregated into a four-way tensor $\boldsymbol{Y}$ of
size $N \times N \times A \times T$, where $N$ is the number of
country actors and $A$ is the number of action types, by aggregating
the events into $T$ time steps on the basis of their timestamps. Each
element $y_{ijat}$ of $\boldsymbol{Y}$ is a count of the number of
actions of type $a$ taken by country $i$ toward country $j$ during
time step $t$. As described in section~\ref{sec:exploratory}, we
experimented with various date ranges and time step granularities. For
example, in one set of experiments, we used the entire ICEWS data set,
spanning 1995 through 2012 (i.e., 18 years) with monthly time
steps---i.e., a $249 \times 249 \times 20 \times 216$ tensor with
267,844,320 elements.

Tensors derived from ICEWS and GDELT are very sparse. For the $249
\times 249 \times 20 \times 216$ ICEWS tensor described above, only
0.54\% of the elements (roughly 1.5 million elements) are
non-zero. Moreover, these non-zero counts are highly dispersed with a
variance-to-mean ratio (VMR) of 57. Any realistic model of such data
must therefore be robust to sparsity and capable of representing
high levels of dispersion.

% TODO: Explain overdispersion and what it means in more depth and why
% it is desirable for a model to be able to express overdispersion.

\section{Bayesian Poisson Tensor\\ Factorization}
\label{sec:BPTF}

Tensor factorization methods decompose an observed $M$-way tensor
$\boldsymbol{Y}$ into $M$ latent factor matrices $\Theta^{(1)},
\ldots, \Theta^{(M)}$ that provide a low-dimensional representation of
the salient patterns in $\boldsymbol{Y}$. There are many different
tensor factorization methods; the two most common methods are the
Tucker decomposition~\cite{tucker66some} and the Canonical Polyadic
(CP) decomposition~\cite{harshman70foundations}. These methods can
both be viewed as tensor generalizations of singular value
decomposition. Here, we focus on the CP decomposition, as it performs
better than the Tucker decomposition when modeling sparse count
data~\cite{kolda08scalable}.

For a four-way count tensor $\boldsymbol{Y}$ of size $N \times N
\times A \times T$, the CP decomposition treats each observed count
$y_{ijat}$ as
\begin{equation}
  \label{eqn:reconstruction}
  y_{ijat} \approx \hat{y}_{ijat} \equiv \sum_{k=1}^K \theta^{(1)}_{ik}
  \theta^{(2)}_{jk} \theta^{(3)}_{ak} \theta^{(4)}_{tk}
  \end{equation}
for $i,j \in [N]$, $a \in [A]$, and $t \in [T]$, where
$\theta^{(1)}_{ik}$, $\theta^{(2)}_{jk}$, $\theta^{(3)}_{ak}$, and
$\theta^{(4)}_{tk}$ are known as \emph{factors}, $\hat{y}_{ijat}$ is
known as the \emph{reconstruction} of count $y_{ijat}$, and
$\hat{\boldsymbol{Y}}$ is the reconstruction of the entire tensor
$\boldsymbol{Y}$. The set of all factors used to model
$\boldsymbol{Y}$ can be aggregated into four latent \emph{factor
  matrices}; for example, $\Theta^{(1)} \equiv \big( \big(
\theta^{(1)}_{ik} \big)_{i=1}^N \big)_{k=1}^K$---an $N \times K$
matrix. Since each factor matrix has $K$ columns, a single index $k
\in [K]$ indexes four columns (one per matrix). These columns are
collectively known as a \emph{component}; for example, component $k$
consists of $\big( \theta^{(1)}_{ik} \big)_{i=1}^N$, $\big(
\theta^{(2)}_{jk}\big)_{j=1}^N$, $\big(
\theta_{ak}^{(3)}\big)_{a=1}^A$, and $\big(
\theta_{tk}^{(4)}\big)_{t=1}^T$---i.e., a length-$N$ vector of sender
factors, a length-$N$ vector of receiver factors, a length-$A$ vector
of action-type factors, and a length-$T$ vector of time-step
factors. Figure 1 visually depicts two components
inferred from ICEWS and GDELT.

When viewed from a probabilistic perspective, the reconstruction
$\hat{y}_{ijat} \equiv \sum_{k=1}^K \theta_{ik}^{(1)}
\theta_{jk}^{(2)} \theta_{ak}^{(3)} \theta_{tk}^{(4)}$ can be thought
of as the mean of the distribution from which the observed count
$y_{ijat}$ is assumed to have been drawn. If this distribution is a
Poisson---i.e., if $y_{ijat} \sim \textrm{Pois}(y_{ijat};
\hat{y}_{ijat})$---then the process of decomposing $\boldsymbol{Y}$
into its latent factor matrices is known as Poisson tensor
factorization (PTF), and can be performed via maximum likelihood
estimation (MLE) of $\Theta^{(1)}$, $\Theta^{(2)}$, $\Theta^{(3)}$,
and $\Theta^{(4)}$. For sparse count data, PTF often yields better
estimates of the latent factor matrices than those obtained by
assuming each count to have been drawn from a Gaussian
distribution---i.e., $y_{ijat} \sim \mathcal{N}(y_{ijat};
\hat{y}_{ijat}, \sigma^2)$~\cite{chi12tensors}.

In this paper, we also assume that each observed count $y_{ijat}$ is
drawn from a Poisson distribution with mean $\hat{y}_{ijat}$; however,
rather than obtaining point estimates of the factor matrices using
maximum likelihood estimation, we impose prior distributions on the
latent factors and perform full Bayesian inference. Bayesian inference
for Poisson matrix factorization (PMF) was originally proposed by
Cemgil~\cite{cemgil09bayesian} and has been successfully used for
several tasks including image reconstruction~\cite{cemgil09bayesian},
music tagging~\cite{liang15codebook-based}, topic
modeling~\cite{paisley14bayesian}, content
recommendation~\cite{gopalan15scalable}, and community
detection~\cite{gopalan13efficient}; here, we generalize Bayesian PMF
to tensors.\footnote{Beyza and Cemgil~\cite{ermis14bayesian} described
  the same model in a paper written concurrently to a previous
  version~\cite{schein14inferring} of this paper.}

Since the Gamma distribution is the conjugate prior for a Poisson
likelihood, Bayesian PMF typically imposes Gamma priors on the latent
factors~\cite{cemgil09bayesian,gopalan15scalable,zhou11beta-negative}. The
Gamma distribution, which has support on $(0, \infty)$, is
  parameterized by a shape parameter $a > 0$ and a rate parameter $b >
  0$; if $\theta \sim \textrm{Gamma}\,(\theta; a, b)$, then
  $\mathbb{E}[\theta] = \frac{a}{b}$ and $\textrm{Var}[\theta] =
  \frac{a}{b^2}$. Thus, when $a \ll 1$ and $b$ is small, the Gamma
  distribution concentrates most of its mass near zero yet maintains a
  heavy tail and can therefore be used as a sparsity-inducing
  prior~\cite{cemgil09bayesian,gopalan15scalable}. We show the effects of
  different $a$ and $b$ values in figure~2.

\begin{figure}[t]
  \label{fig:Gamma}
  \begin{center}
    \includegraphics[width=0.75\linewidth]{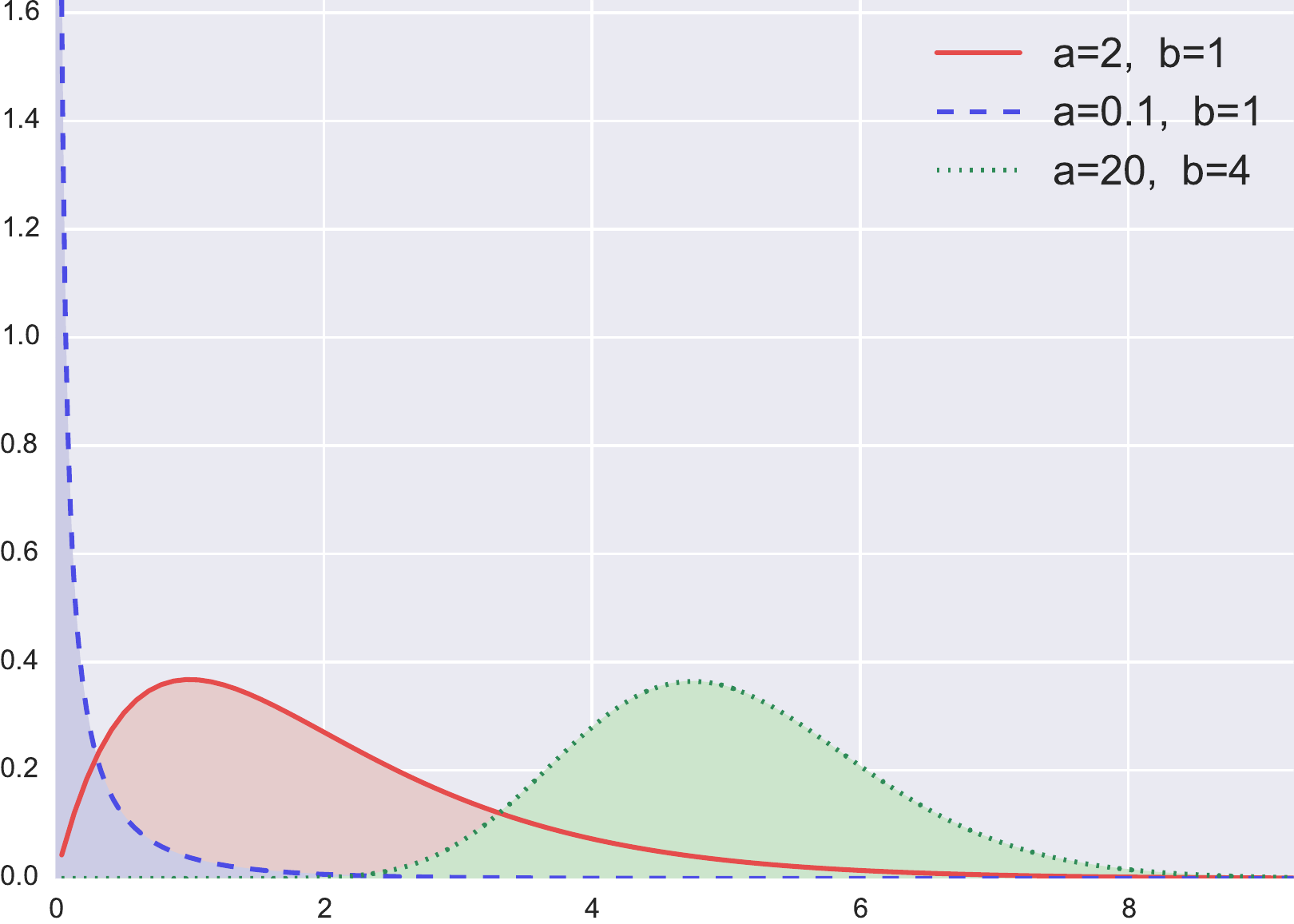}
  \end{center}
  \vspace{-1em}
  \caption{\scriptsize Three Gamma distributions with different values
    for the shape $a$ and rate $b$ parameters. The distribution induces
    sparsity when $a \ll 1$ and $b$ is small (shown in blue).}
  \vspace{-1em}
\end{figure}

To define Bayesian Poisson tensor factorization (BPTF) for a four-way
tensor, we impose four sparsity-inducing Gamma priors over the latent
factors.  For a single factor, e.g., $\theta^{(1)}_{ik}$,
\begin{equation}
  \label{eq:q_for_single_theta}
  \theta^{(1)}_{ik} \sim \textrm{Gamma}\,(\theta^{(1)}_{ik}; \alpha,
  \alpha\, \beta^{(1)}),
  \end{equation}
and similarly for $\theta^{(2)}_{jk}$, $\theta^{(3)}_{ak}$, and
$\theta^{(4)}_{tk}$. Under this parameterization of the Gamma
distribution, where the rate parameter is the product of the shape
parameter and $\beta^{(1)}$, the mean of the prior is completely
determined by $\beta^{(1)}$ (since $\mathbb{E}[\theta_{ik}^{(1)}] =
\frac{\alpha}{\alpha \beta^{(1)}} = \frac{1}{\beta^{(1)}}$), which can be inferred from the data~\cite{cemgil09bayesian,liang15codebook-based}. The shape
parameter $\alpha$, which determines the sparsity of the latent factor
matrices, can be set by the user. Throughout our experiments, we use
$\alpha = 0.1$ to encourage sparsity and hence promote
interpretability of the factors.

% TODO: Add point about how the Gamma-Poisson expresses overdispersion

\section{Variational inference}
\label{sec:variational}

Given an observed tensor $\boldsymbol{Y}$, Bayesian inference of the
latent factors involves ``inverting'' the generative process described
in the previous section to obtain the posterior distribution of the
latent factor matrices conditioned on $\boldsymbol{Y}$ and the model
hyperparameters $\mathcal{H} \equiv \{\alpha, \beta^{(1)},
\beta^{(2)}, \beta^{(3)}, \beta^{(4)} \}$:
\begin{displaymath}
P\left(\Theta^{(1)},\Theta^{(2)},\Theta^{(3)},\Theta^{(4)} \,|\, \mathbf{Y}, \mathcal{H}\right).
\end{displaymath}

The posterior distribution for BPTF is analytically intractable and
must be approximated. Variational inference turns the process of
approximating the posterior distribution into an optimization
algorithm. It involves first specifying a parametric family of
distributions $Q$ over the latent variables of interest, indexed by
the values of a set of \emph{variational parameters} $\mathcal{S}$.
The functional form of $Q$ is typically chosen so as to facilitate
efficient optimization of $\mathcal{S}$.  Here, we use a fully
factorized \emph{mean-field approximation} and define $Q$ to be the
product of $N \cdot N \cdot A \cdot T \cdot K$ independent Gamma
distributions---one for each latent factor---e.g., for
$\theta_{ik}^{(1)}$,
\begin{equation}
Q(\theta^{(1)}_{ik}; \mathcal{S}^{(1)}_{ik}) =
\textrm{Gamma}\,(\theta^{(1)}_{ik}; \gamma^{(1)}_{ik},
\delta^{(1)}_{ik}),
\end{equation}
where $\mathcal{S}^{(1)} \equiv \big(\big(\gamma^{(1)}_{ik},
\delta^{(1)}_{ik}\big)_{i=1}^N\big)_{k=1}^K$. The full set of
variational parameters is thus $\mathcal{S} \equiv
\{\mathcal{S}^{(1)}, \mathcal{S}^{(2)}, \mathcal{S}^{(3)},
\mathcal{S}^{(4)}\}$. This form of $Q$ is similar to that used in
Bayesian PMF~\cite{cemgil09bayesian, paisley14bayesian,
  gopalan15scalable}.

The variational parameters are then fit so as to yield the closest
member of $Q$ to the exact posterior---known as the \emph{variational
  distribution}. Specifically, the algorithm sets the values of
$\mathcal{S}$ to those that minimize the KL divergence of the exact
posterior from $Q$. It can be shown that these values are the same as
those that maximize a lower bound on $P(\boldsymbol{Y} \,|\,
\mathcal{H})$, known as the \emph{evidence lower bound} (ELBO):
\begin{equation*}
  \mathcal{B}(\mathcal{S}) = \mathbb{E}_{Q}\left[
    \log{\left(P(\boldsymbol{Y}, \Theta^{(1)}, \Theta^{(2)},
      \Theta^{(3)}, \Theta^{(4)} \,|\, \mathcal{H})\right)}\right] +
  H(Q),
  \end{equation*}
where $H(Q)$ is the entropy of $Q$. When $Q$ is a fully factorized
approximation, finding values of $\mathcal{S}$ that maximize the ELBO
can be achieved by performing coordinate ascent, iteratively updating
each variational parameter, while holding the others fixed, until
convergence (defined by relative change in the ELBO).  The update
equation for each parameter can be derived easily using an auxiliary
variable as shown
for Bayesian PMF~\cite{cemgil09bayesian,
  paisley14bayesian,gopalan15scalable}; we therefore
omit derivations.

For parameters $\gamma^{(1)}_{ik}$ and $\delta^{(1)}_{ik}$, the update
equations are
\begin{align}
  \label{eqn:gamma_update}
\gamma^{(1)}_{ik} &:= \alpha + \sum_{j,a,t} y_{ijat}
\frac{\mathbb{G}_Q\left[
    \theta^{(1)}_{ik} \theta^{(2)}_{jk}
    \theta^{(3)}_{ak} \theta^{(4)}_{tk}
    \right]}{\sum_{k=1}^K
  \mathbb{G}_Q\left[
  \theta^{(1)}_{ik} \theta^{(2)}_{jk} \theta^{(3)}_{ak}
  \theta^{(4)}_{tk} \right]}\\
\label{eqn:delta_update}
\delta^{(1)}_{ik} &:= \alpha\,\beta^{(1)} + \sum_{j,a,t}
\mathbb{E}_Q\left[ \theta^{(2)}_{jk}
  \theta^{(3)}_{ak} \theta^{(4)}_{tk} \right],
\end{align}
where $\mathbb{E}_Q \left[ \cdot \right]$ and $\mathbb{G}_Q \left[
  \cdot \right] = \exp{( \mathbb{E}_Q \left[ \log{( \cdot )}\right]
  )}$ denote arithmetic and geometric expectations. Since $Q$ is fully
factorized, each expectation of a product can be factorized into a
product of individual expectations, which, e.g., for
$\theta_{ik}^{(1)}$ are
\begin{equation}
  \mathbb{E}_Q \left[ \theta^{(1)}_{ik}\right] =
  \frac{\gamma^{(1)}_{ik}}{\delta^{(1)}_{ik}} \quad\textrm{and}\quad
  \mathbb{G}_Q \left[ \theta^{(1)}_{ik} \right]  = \frac{\exp \big(
      \Psi \big( \gamma^{(1)}_{ik} \big) \big)}{\delta^{(1)}_{ik}},
  \end{equation}
where $\Psi(\cdot)$ is the digamma function. Each expectation---a
sufficient statistic---can be cached to improve efficiency. Note that
the summand in (\ref{eqn:gamma_update}) need only be computed for
those values of $j$, $a$, and $t$ for which $y_{ijat} > 0$; provided
$\boldsymbol{Y}$ is very sparse, inference is efficient even for very
large tensors.

The hyperparameters $\beta^{(1)}$, $\beta^{(2)}$, $\beta^{(3)}$, and
$\beta^{(4)}$ can be optimized via an empirical Bayes method, in which
each hyperparameter is iteratively updated along with the variational
parameters according to the following update equation:
\begin{equation}
  \label{eqn:beta_update}
\beta^{(1)} := \left(\sum_{i,k} \mathbb{E}_Q\left[ \theta_{ik}^{(1)} \right]\right)^{-1}.
\end{equation}

Update equations~(\ref{eqn:gamma_update}), (\ref{eqn:delta_update}),
and (\ref{eqn:beta_update}) completely specify the variational
inference algorithm for BPTF. Our Python implementation, which is
intended to support arbitrary $M$-way tensors in addition to the
four-way tensors described in this paper, is available for use under
an open source license\footnote{\url{https://github.com/aschein/bptf}}.

\section{Predictive Analysis}
\label{sec:eval}

\begin{table*}
\label{fig:predict}
\scriptsize
\caption{\scriptsize Out-of-sample predictive performance for our
  model (BPTF) and non-negative tensor factorization with Euclidean
  distance (NTF-LS) and generalized KL divergence (NTF-KL or,
  equivalently, PTF). Each row contains the results of a single
  experiment.  ``I-top-25'' means the experiment used data from ICEWS
  and we predicted the upper-left $25 \times 25$ portion of each test
  slice (and treated its complement as observed). ``G-top-100$^{c}$''
  means the experiment used data from GDELT and we predicted the
  complement of the upper-left $100 \times 100$ portion of each test
  slice. For each experiment, we state the density (percentage of
  non-zero elements) and VMR (i.e., dispersion) of the unobserved
  portion of the test set. We report three types of error: mean
  absolute error (MAE), mean absolute error on non-zero elements
  (MAE-NZ), and Hamming loss on the zero elements (HAM-Z). All models
  achieved comparable scores when we predicted the sparser portion of
  each test slice (bottom four rows). BPTF significantly outperformed
  the other models when we predicted the denser $25 \times 25$ or $100
  \times 100$ portion (top four rows).}
\begin{center}
\begin{tabular}{l l l l l l l l l l l l} \toprule
 & &  & \multicolumn{3}{c}{NTF-LS} & \multicolumn{3}{c}{NTF-KL (PTF)} & \multicolumn{3}{c}{BPTF}\\ \cmidrule(r){4-6} \cmidrule(r){7-9} \cmidrule(r){10-12}
 & Density & VMR & MAE & MAE-NZ & HAM-Z & MAE & MAE-NZ & HAM-Z & MAE & MAE-NZ & HAM-Z\\ \midrule
I-top-25 & 0.1217 & 105.8755 &34.4 & 217 & 0.271 & 8.37 & 56.7 & 0.138 & {\bf 1.99} & {\bf 12.9} & {\bf 0.113}\\ %\midrule
G-top-25 & 0.2638 & 180.4143 &52.5 & 167 & 0.549 & 15.5 & 53.7 & 0.327 & {\bf 8.94} & {\bf 29.8} & {\bf 0.292}\\ %\midrule
I-top-100 & 0.0264 & 63.1118 &29.8 & 979 & 0.0792 & 10.5 & 346 & 0.0333 & {\bf 0.178} & {\bf 5.05} & {\bf 0.0142}\\ %\midrule
G-top-100 & 0.0588 & 111.8676 &42.6 & 470 & 0.217 & 4 & 58.6 & 0.0926 & {\bf 0.95} & {\bf 12.2} & {\bf 0.0682}\\ \midrule
I-top-25$^{c}$ & 0.0021 & 8.6302 &{\bf 0.00657} & {\bf 2.27} & {\bf 0.00023} & 0.0148 & 2.72 & 0.00256 & 0.0104 & 2.31 & 0.00161\\ %\midrule
G-top-25$^{c}$ & 0.0060 & 20.4858 &0.0435 & 4.4 & {\bf 0.00474} & 0.0606 & 4.9 & 0.00893 & {\bf 0.0412} & {\bf 4.01} & 0.00601\\ %\midrule
I-top-100$^{c}$ & 0.0004 & 4.4570 &{\bf 0.000685} & 1.63 & {\bf 3.33e-07} & 0.0011 & {\bf 1.55} & 5.43e-05 & 0.00109 & 1.56 & 4.97e-05\\ %\midrule
G-top-100$^{c}$ & 0.0015 & 9.9432 &{\bf 0.00584} & 3.23 & {\bf 0.000112} & 0.0084 & {\bf 2.97} & 0.00109 & 0.00803 & 3 & 0.000957\\ %\midrule
\bottomrule
\end{tabular}
\end{center}
\end{table*}

We validated our model by comparing its predictive performance to that
of standard methods for non-negative tensor factorization involving
maximum likelihood estimation.

\vspace{0.5em}

\noindent\textbf{Baselines:\,} Non-Bayesian methods for CP
decomposition find values of the latent factor matrices that minimize
some cost function of the observed tensor $\boldsymbol{Y}$ and its
reconstruction $\hat{\boldsymbol{Y}}$. Researchers have proposed many
cost functions, but most often use Euclidean distance or generalized
KL divergence, preferring the latter when the observed tensor consists
of sparse counts. Generalized KL divergence is
\begin{equation}
  \label{eq:kl}
  D( \boldsymbol{Y}\,||\, \hat{\boldsymbol{Y}}) = - \sum_{i,j,a,t}
  \left( y_{ijat} \log{(\hat{y}_{ijat})} - \hat{y}_{ijat} \right) + C,
  \end{equation}
where constant $C \equiv \sum_{i,j,a,t} \left( y_{ijat}
\log{(y_{ijat})} - y_{ijat} \right)$ depends on the observed
data only. The standard method for estimating the values of the latent
factors involves multiplicative update equations, originally
introduced for matrix factorization by Lee and
Seung~\cite{lee99learning} and later generalized to tensors by Welling
and Weber~\cite{welling01positive}. The multiplicative nature of these
update equations acts as a non-negativity constraint on the factors
which promotes interpretability and gives the algorithm its name:
non-negative tensor factorization (NTF).

\begin{figure}[t]
\label{fig:heat}
\begin{center}
    \includegraphics[width=\linewidth]{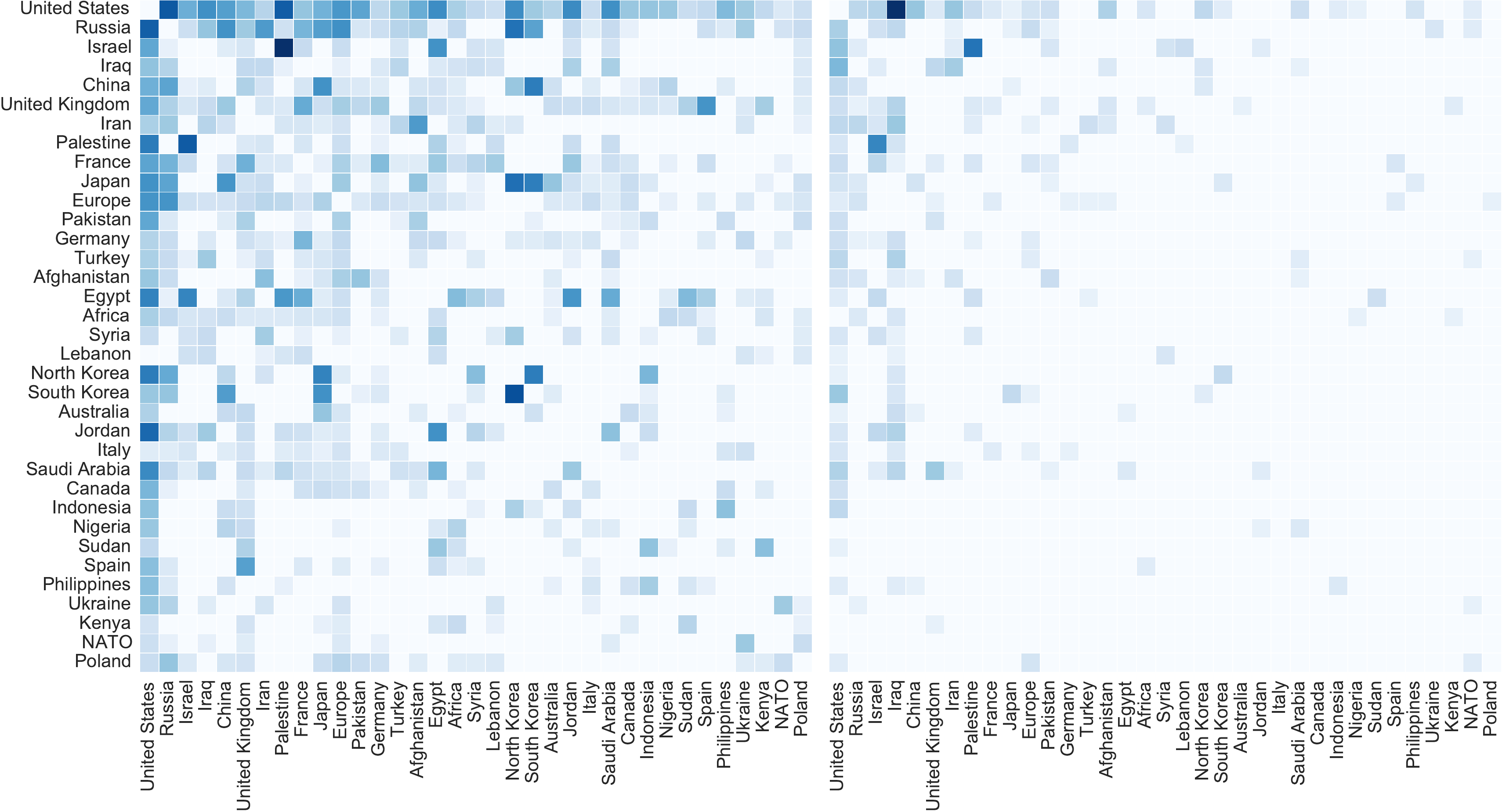}
  \end{center}
\vspace{-1em}
\caption{\scriptsize Sender--receiver slices from the GDELT tensor
  spanning 1990 through 2007, with monthly time steps (i.e.,
  $T=216$). Both slices correspond to $t=151$ (July 2002).  The left
  slice corresponds to \emph{Intend to Cooperate}, while the right
  slice corresponds to \emph{Threaten}. We sorted the country actors
  by their overall activity so that the slices were generally denser
  toward the upper-left corner; only the upper-left $35 \times 35$
  portion of each slice is shown here. The three darkest elements
  (i.e., highest counts) in the second slice correspond to Israel
  $\rightarrow$ Palestine, Palestine $\rightarrow$ Israel, and US
  $\rightarrow$ Iraq.}
\vspace{-1em}
\end{figure}

Some cost functions also permit a probabilistic interpretation:
finding values of the latent factors that minimize them is equivalent
to maximum likelihood estimation of a probabilistic model. The log
likelihood function of a Poisson tensor factorization model---$y_{ijat}
\sim \textrm{Pois}(y_{ijat}; \hat{y}_{ijat})$---is
\begin{align}
\mathcal{L}(\hat{\boldsymbol{Y}}; \boldsymbol{Y}) &=
\log{\left(\prod_{i,j,a,t}
  \frac{(\hat{y}_{ijat})^{y_{ijat}}}{y_{ijat}!}
  \exp{\left(-\hat{y}_{ijat}\right)} \right)}\\
\label{eq:ll}
&= \sum_{i,j,a,t} \left( y_{ijat} \log{(\hat{y}_{ijat})} - \hat{y}_{ijat} \right) + C,
\end{align}
where constant $C \equiv \sum_{i,j,a,t} -\log{\left(
  y_{ijat}!\right)}$ depends on the observed data only. Since
equation~(\ref{eq:kl}) is equal to the negative of equation
(\ref{eq:ll}) up to a constant, maximum likelihood estimation for
Poisson tensor factorization is equivalent to minimizing the
generalized KL divergence of $\hat{\boldsymbol{Y}}$ from
$\boldsymbol{Y}$.

To validate our modeling assumptions, we compared the out-of-sample
predictive performance of BPTF to that of non-negative tensor
factorization with Euclidean distance (NTF-LS) and non-negative tensor
factorization with generalized KL divergence (NTF-KL or, equivalently
PTF).

\vspace{0.5em}

\noindent\textbf{Experimental design:\,} Using both ICEWS and GDELT,
we explored how well each model generalizes to out-of-sample data with
varying degrees of sparsity and dispersion. For each data set---ICEWS
or GDELT---we sorted the country actors by their overall activity (as
both sender and receiver) so that the $N \times N$ sender--receiver
slices of the observed tensor were denser toward the upper-left
corner. Figure 3 depicts this property. We then divided
the observed tensor into a training set and a test set by randomly
constructing an 80\%--20\% split of the time steps. We defined
training set $\boldsymbol{Y}^{\text{train}}$ to be the $N \times N
\times A$ slices of $\boldsymbol{Y}$ indexed by the time steps in the
80\% split and defined test set $\boldsymbol{Y}^{\textrm{test}}$ to be
the $N \times N \times A$ slices indexed by the time steps in the 20\%
split.

We compared the models' predictive performance in two scenarios,
intended to test their abilities to handle different levels of
sparsity and dispersion: one in which we treated the denser upper-left $N' \times
N'$ (for some $N' < N$) portion of each test slice as
observed at test time and predicted its complement, and one in which
we observed the complement at test time and predicted the denser $N'
\times N'$ portion.

In each setting, we used an experimental strategy analogous to
\emph{strong generalization} for collaborative
filtering~\cite{marlin04collaborative}. During training, we fit each
model to the fully observed training set. We then fixed the values of
the variational parameters for the sender, receiver, and action-type
factor matrices (or direct point estimates of the factors,
for the non-Bayesian models) to those inferred from the training set. For each test
slice, indexed by time step $t$, we used the observed upper-left $N'
\times N'$ portion (or its complement) to infer variational parameters
for (or direct point estimates of) its time-step factors $\{
\theta_{tk}^{(4)} \}_{k=1}^K$. Finally, we reconstructed the missing
portion of each test slice using
equation~(\ref{eqn:reconstruction}). For the reconstruction step, we
can 
obtain point estimates of the latent factors by
taking their arithmetic expectations or their geometric expectations under the
variational distribution. In this section, we report results obtained
using geometric expectations only; we explain this
choice in section~\ref{sec:discussion}.

\begin{figure*}[ht]
\label{fig:found_in_both} 
  \begin{center}
    \includegraphics[width=0.49\linewidth]{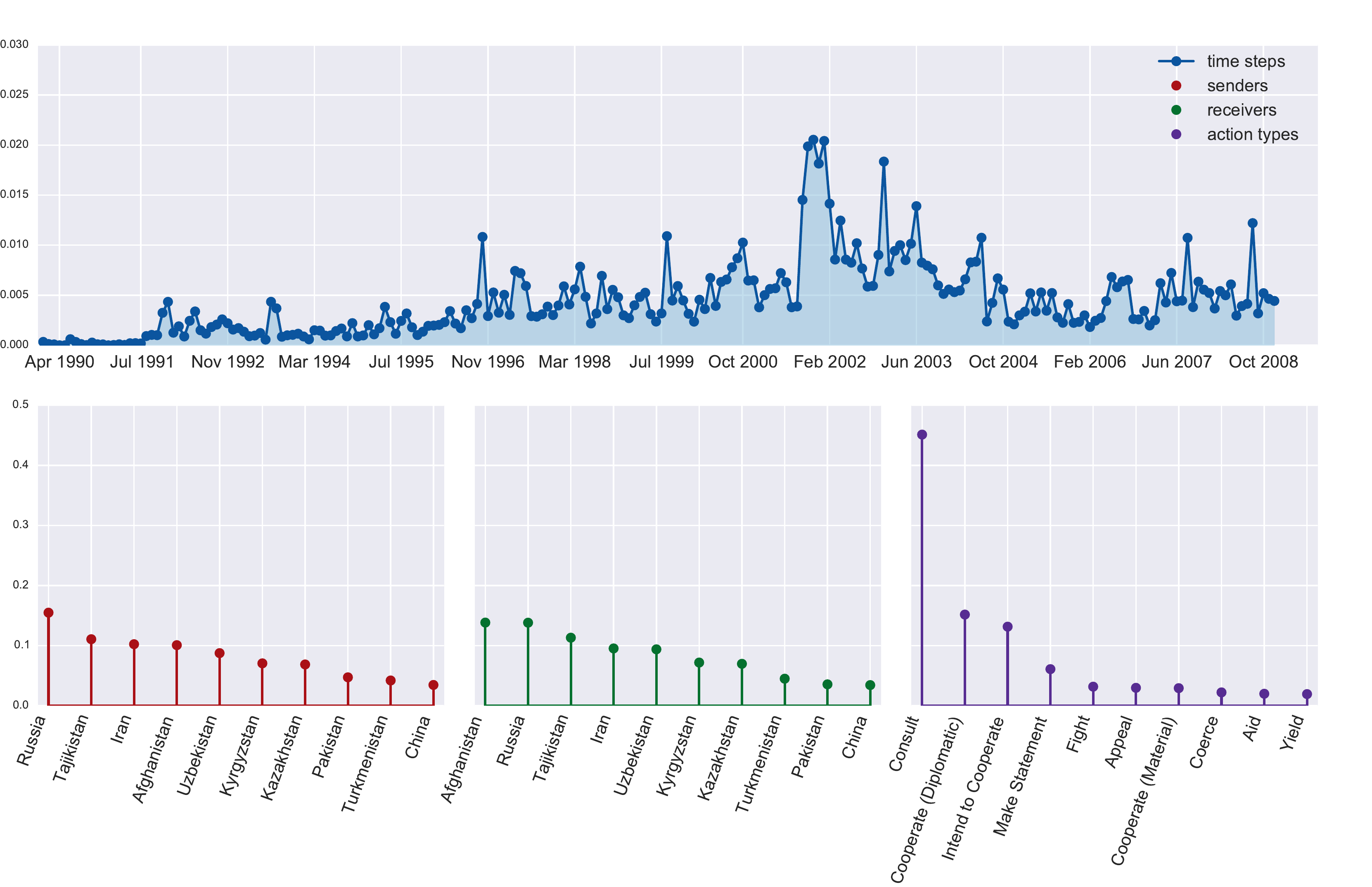}
    \includegraphics[width=0.49\linewidth]{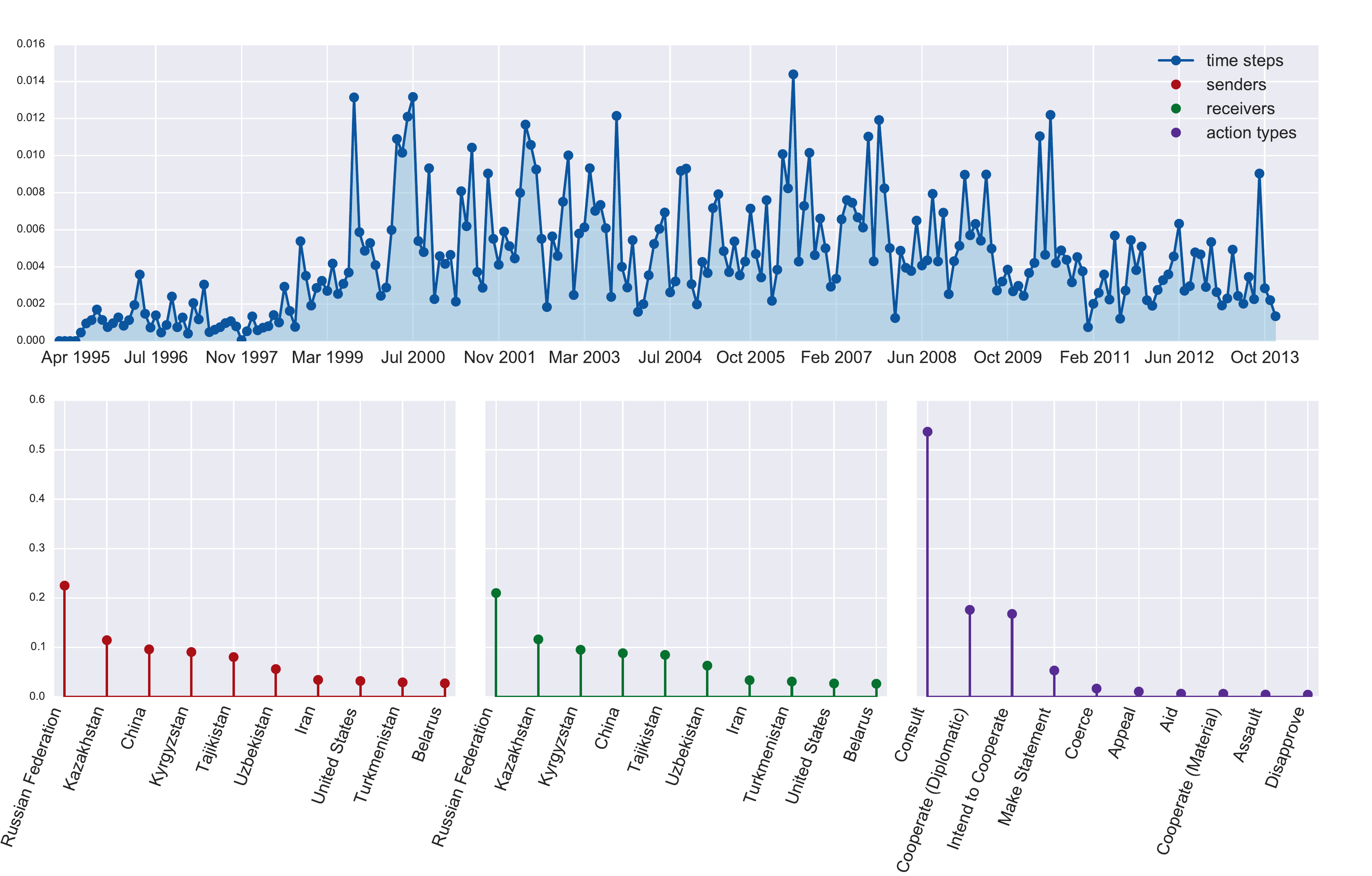}
  \end{center}
  \vspace{-2em}  
  \caption{\scriptsize Regional relations between Central Asian
    republics and regional superpowers, found in both GDELT (left;
    spanning 1990 through 2007, with monthly time steps) and ICEWS
    (right; spanning 1995 through 2012, with monthly time steps).  }
  \vspace{-1em}
\end{figure*}

The time-step factors inferred from the observed portion of a given
test slice capture the extent to which the sender, receiver, and
action-type factors for each component inferred from the training set
describe (the observed portion of) that slice. For example, if
component $k$ summarizes the Israeli--Palestinian conflict, with
Israel and Palestine as top actors and \emph{Fight} as a top action
type, then if Israeli--Palestinian hostilities were intense during
test time step $t$ and if Israel and Palestine belong to the observed
portion of each test slice, the inferred value of $\theta^{(4)}_{tk}$
is very likely to be large.

We used the entire ICEWS data set from 1995 through 2012 (i.e., 18
years), with events aggregated into monthly time steps. The resultant
tensor was of size $249 \times 249 \times 20 \times 216$. Since GDELT
covers a larger date range (1979 to the present) than ICEWS, we
therefore selected an 18-year subset of GDELT spanning
1990 through 2007, and aggregated events into monthly time steps to
yield a tensor of size $223 \times 223 \times 20 \times 216$. Since we
are interested in interactions between countries, we omitted
self-actions so that the diagonal of each $N \times N$
sender--receiver slice was zero. Ranking the country actors by their
overall activity (as both sender and receiver), the top four actors in
the ICEWS tensor are USA, Russia, China, and Israel, while the top
four actors in the GDELT tensor are USA, Russia, Israel, and Iraq. The
GDELT tensor contains many more events than the ICEWS tensor (26
million events versus six million events). It is also much denser
(1.6\% of the elements are non-zero, as opposed to 0.54\%) and
exhibits a much higher level of dispersion (VMR of 100, as opposed to
57).

\vspace{0.5em}

\noindent\textbf{Summary of results:\,} The out-of-sample predictive performance of each model is shown in
table~1. We experimented with several different values
of $K$ and found that all three models were insensitive to its value;
we therefore report only those results obtained using $K=50$. We
computed three types of error: mean absolute error (MAE), mean
absolute error on only non-zero elements (MAE-NZ), and Hamming loss on
only the zero elements (HAM-Z). HAM-Z corresponds to the fraction of
true zeros in the unobserved portion of the test set (i.e., elements
for which $y_{ijat} = 0$) whose reconstructions were (incorrectly)
predicted as being greater than 0.5. For each data set, we generated
three training--test splits, and averaged the error scores for each
model across them. For each experiment included in
table~1, we display the density and dispersion of the
corresponding test set. When we treated the dense upper-left $N'
\times N'$ portion as observed at test time (and predicted its
complement), all models performed comparably. In this scenario, NTF-LS
consistently achieved the lowest MAE score and the lowest HAM-Z score,
but not the lowest MAE-NZ score. This pattern suggests that NTF-LS
overfits the sparsity of the training set: when the unobserved portion
of the test set is much sparser than the training set (as it is in
this scenario), NTF-LS achieves lower error scores by simply
predicting many more zeros than NTF-KL (i.e., PTF) or BPTF. In the
opposite scenario, when we observed the complement at test time and
predicted the denser $N' \times N'$ portion, NTF-LS produced
significantly worse predictions than the other models, and our model
(BPTF) achieved the lowest MAE, MAE-NZ, and HAM-Z scores---in some
cases by an order of magnitude over NTF-KL. These results suggest that
in the presence of sparsity, BPTF is a much better model for the
``interesting'' portion of the tensor---i.e., the dense non-zero
portion. This observation is consistent with previous work by Chi and
Kolda which demonstrated that NTF can be unstable, particularly when
the observed tensor is very sparse~\cite{chi12tensors}. In
section~\ref{sec:discussion}, we provide a detailed discussion
comparing NTF and BPTF, and explain why BPTF overcomes the
sparsity-related issues often suffered by NTF.

\section{Exploratory Analysis}
\label{sec:exploratory}

In this section, we focus on the interpretability of the latent
components inferred using our model. (Recall that each latent factor
matrix has $K$ columns; a single index $k \in [K]$ indexes a column in
each matrix---$\big( \theta_{ik}^{(1)} \big)_{i=1}^N$, $\big(
\theta_{jk}^{(2)} \big)_{j=1}^N$, $\big( \theta_{ak}^{(3)}
\big)_{a=1}^A$, and $\big( \theta_{tk}^{(4)}
\big)_{t=1}^T$---collectively known as a component.) We used our model
to explore data from GDELT and ICEWS with several date ranges and time
step granularities, including the 18-year, monthly-time-step tensors
described in the previous section (treated here as fully observed).

When inferring factor matrices from data that span a large date range
(e.g., 18 years), we expect that the inferred components will
correspond to multilateral relations that persist or recur over
time. Figure~1 shows two such components, inferred
from the 18-year GDELT and ICEWS tensors. The first component
corresponds to ongoing negotiations over North Korea's nuclear program,
while the second corresponds to a decade-long war (though precipitated
by a sudden anomalous event). We
found that many the components inferred from 18-year tensors summarize
regional relations---i.e., multilateral relations that persist due to
geographic proximity---similar to those found by
Hoff~\cite{hoff14multilinear}.

\begin{figure*}[ht]
\label{fig:more-anomalies} 
  \begin{center}
    \includegraphics[width=0.49\linewidth]{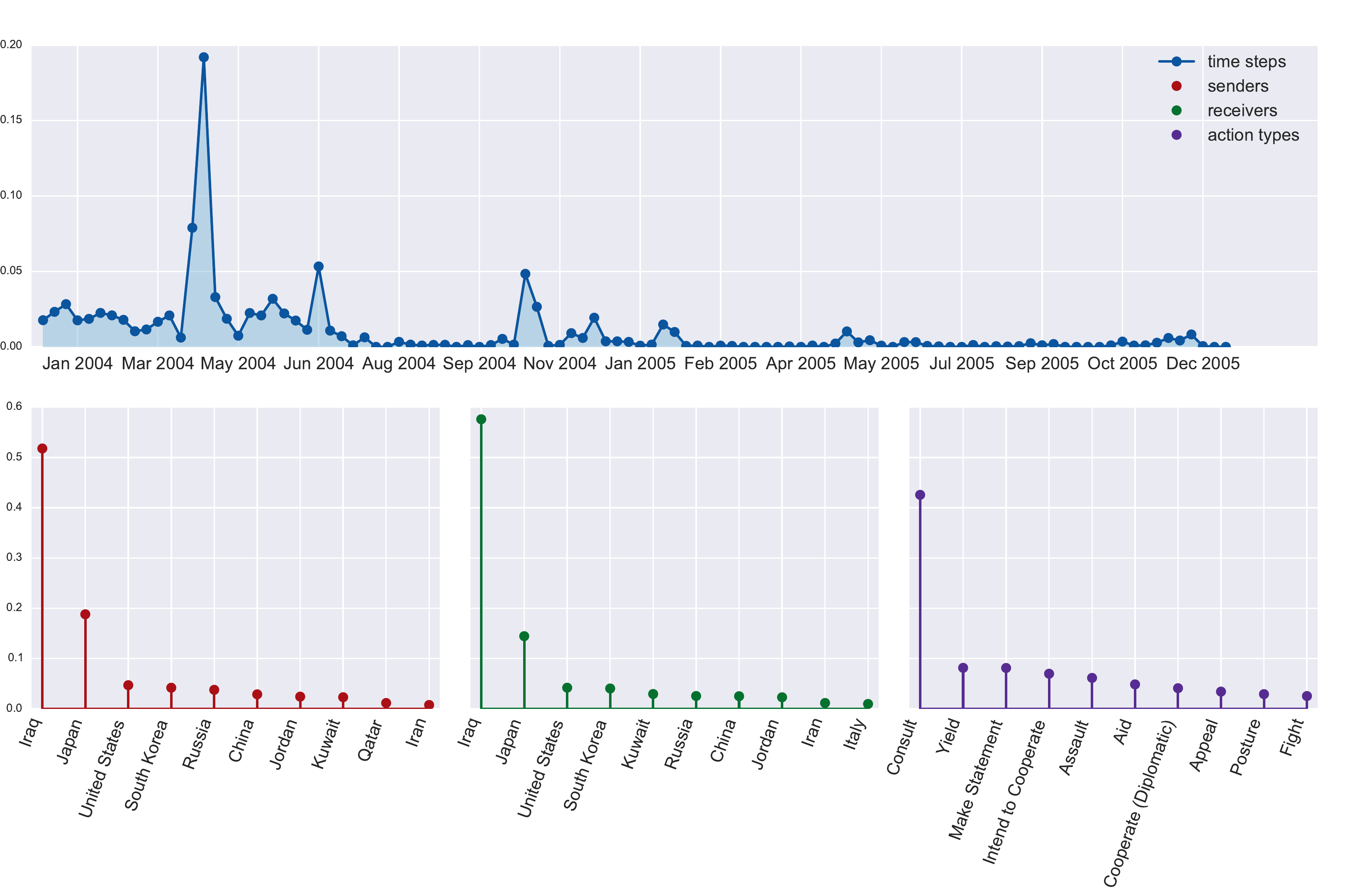}
    \includegraphics[width=0.49\linewidth]{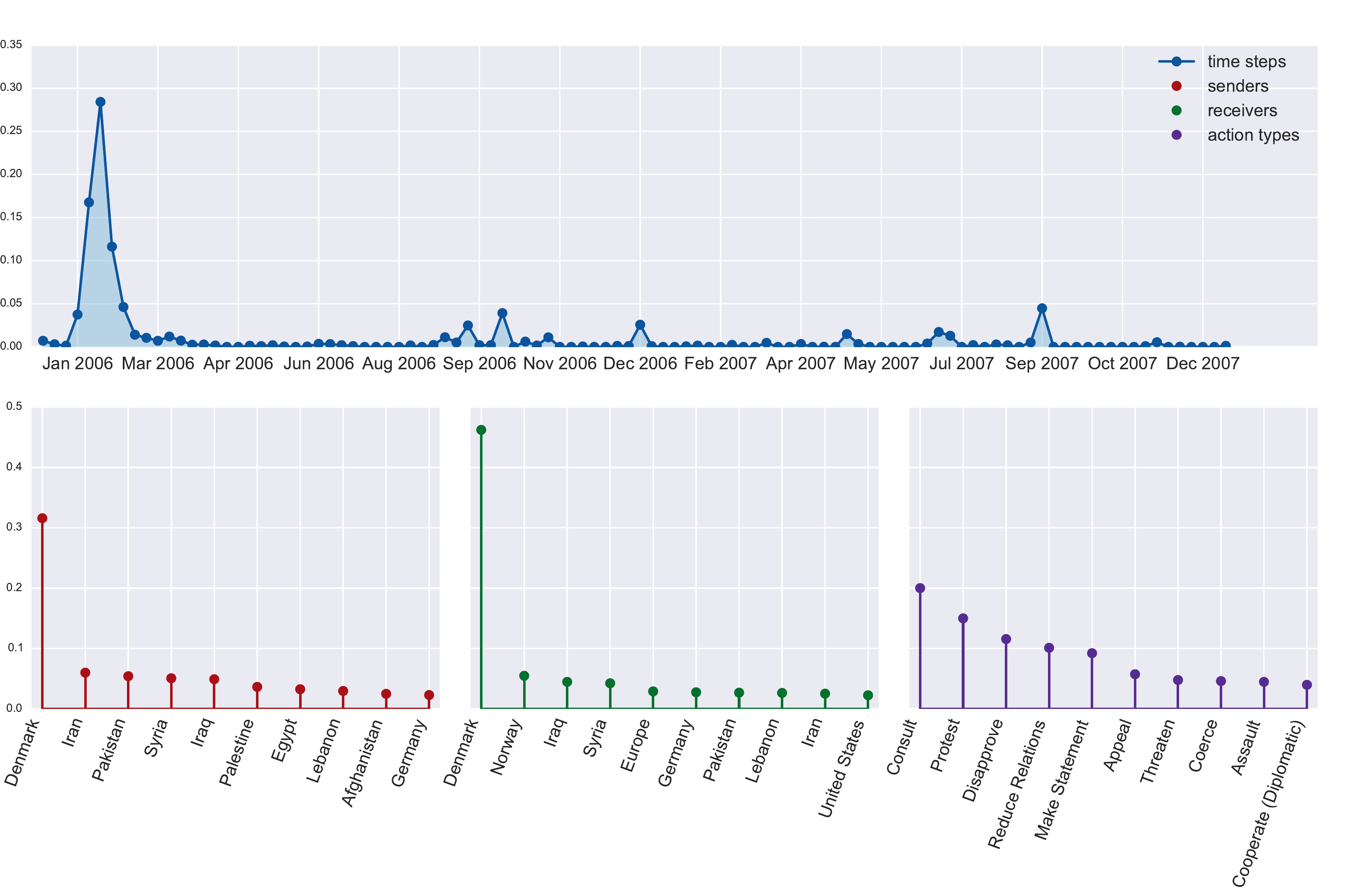}
  \end{center}
  \vspace{-2em}
  \caption{\scriptsize \emph{Left}: Three Japanese citizens were taken
    hostage in Iraq during April 2004 and a third was found murdered
    in October 2004~\cite{wikipedia15japanese}. This component
    inferred from GDELT (2004 through 2005, weekly time steps) had
    the sparsest time-step factor vector. We performed a web search
    for \emph{japan iraq april 2004} to interpret this
    component. \emph{Right}: Protests erupted in Islamic countries
    after a Danish newspaper published cartoons depicting the Prophet
    Muhammad~\cite{wikipedia15jyllands-posten}. Denmark and Iran cut
    diplomatic ties in February 2006 after protesters attacked the
    Danish embassy in Tehran. This component inferred from GDELT
    (2006 through 2007, weekly time steps) had the second sparsest
    time-step factor vector.  Web search: \emph{denmark iran january
      2006}.}
  \vspace{-1em}
\end{figure*}

We found a high correspondence between the regional components
inferred from GDELT and the regional components inferred from ICEWS,
despite the five-year difference in their date
ranges. Figure~4 illustrates this
correspondence. We also found that components summarizing regional
relations exhibited the least sparsity in their sender, receiver, and
time-step factors. For example, the component depicted in
figure~4 has near-uniform values for the top ten
sender and receiver actors (all of whom are regional to Central Asia),
while the time-step factors possess high activity throughout. In
contrast, the time-step factors for the component shown in the second
plot of figure~1 (i.e., the War on Terror) exhibit a
major spike in October 2001. This component's sender and receiver
factors also exhibit uneven activity over the top ten actors, with the
US, Afghanistan, and Pakistan dominating.

These ``regional relations'' components conform to our understanding
of international affairs and foster confidence in BPTF as an
exploratory analysis tool. However, for the same reason, they are also
less interesting. To explore temporally localized multilateral
relations---i.e., anomalous interaction patterns that do not simply
reflect usual activity---we used our model to infer components from
several subsets of GDELT and ICEWS, each spanning a two-year date
range with weekly time steps. We ranked the inferred components by the
sparsity of their time-step factors, measured using the Gini
coefficient~\cite{dorfman79formula}. Ranking components by their Gini
coefficients is a form of \emph{anomaly detection}: components with
high Gini coefficients have unequal time-step factor values---i.e.,
dramatic spikes. Figure~6 shows the highest-ranked
(i.e., most anomalous) component inferred from a subset of GDELT
spanning 2011--2012. This component features an unusual group of top
actors and a clear burst of activity around June 2012. To interpret
this component, we performed a web search for \emph{ecuador UK sweden june
2012} and found that the top hit was a Wikipedia
page~\cite{wikipedia14embassy} about Julian Assange, the
editor-in-chief of the website WikiLeaks---an Australian national,
wanted by the US and Sweden, who sought political asylum at the
Ecuadorian embassy in the UK during June through August 2012. These
countries are indeed the top actors for this component, while the
time-step factors and top action types (i.e., \emph{Consult},
\emph{Aid}, and \emph{Appeal}) track the dates and nature of the
reported events.

In general, we found that when our existing knowledge was insufficient
to interpret an inferred component, performing a web search for the
top two-to-four actors along with the top time step resulted in either
a Wikipedia page or a news article that provided an explanation. We
present further examples of the most anomalous components inferred
from other two-year date ranges in figure~5,
along with the web searches that we performed in order to interpret them.

\begin{figure}[t]
\label{fig:assange}
  \begin{center}
    \includegraphics[width=\linewidth]{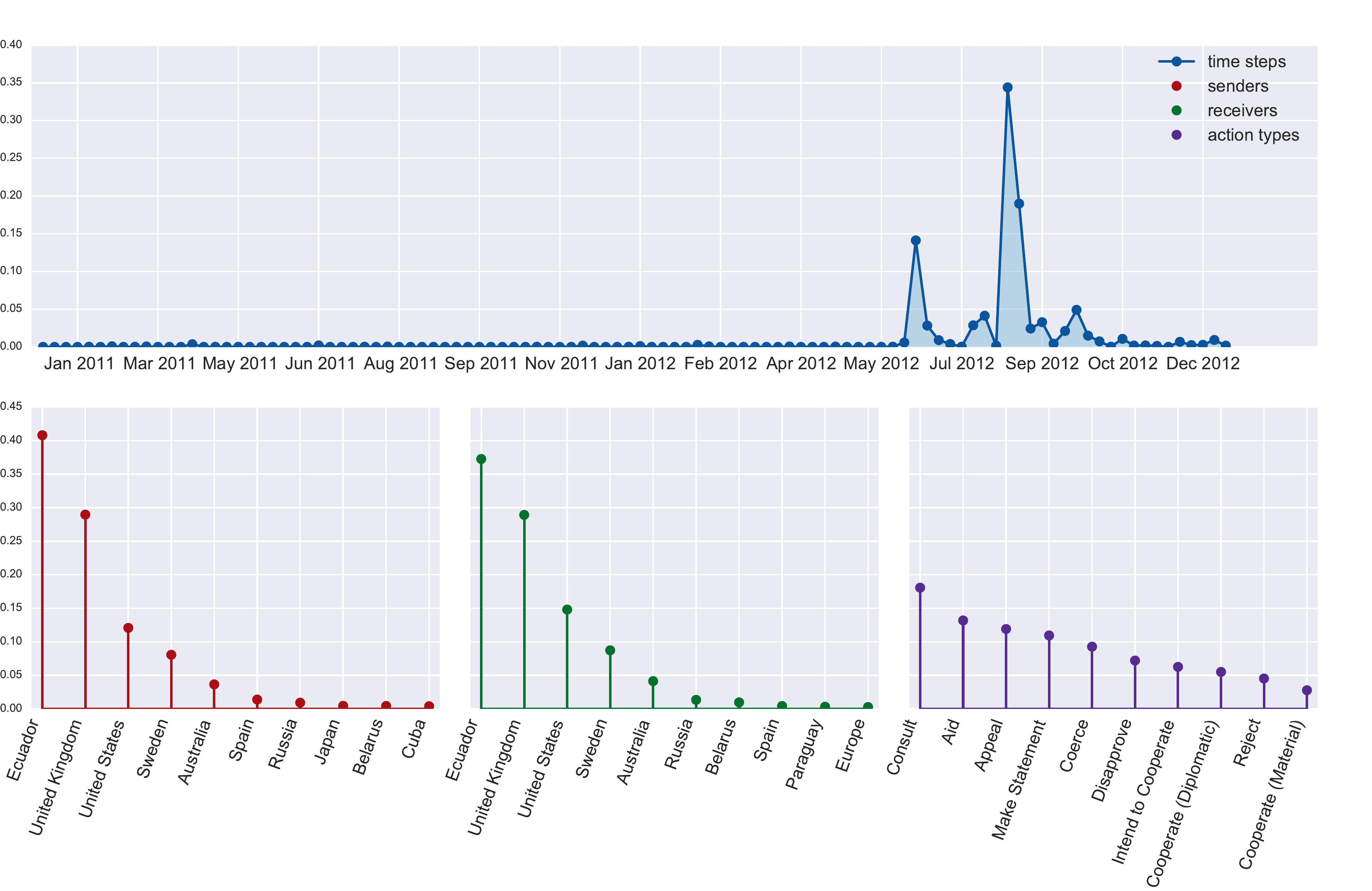}
  \end{center}
  \vspace{-2em}
  \caption{\scriptsize Julian Assange, editor-in-chief of WikiLeaks,
    sought asylum at the Ecuadorian embassy in the UK during June
    through August 2012. This component inferred from GDELT (2011
    through 2012, with weekly time steps) had the sparsest time-step
    factor vector.  We performed a web search for \emph{ecuador UK
      sweden june 2012} to interpret this component.}
  \vspace{-1em}
\end{figure}

\begin{figure*}[ht]
\label{fig:geo} 
  \begin{center}
    \includegraphics[width=1.0\linewidth]{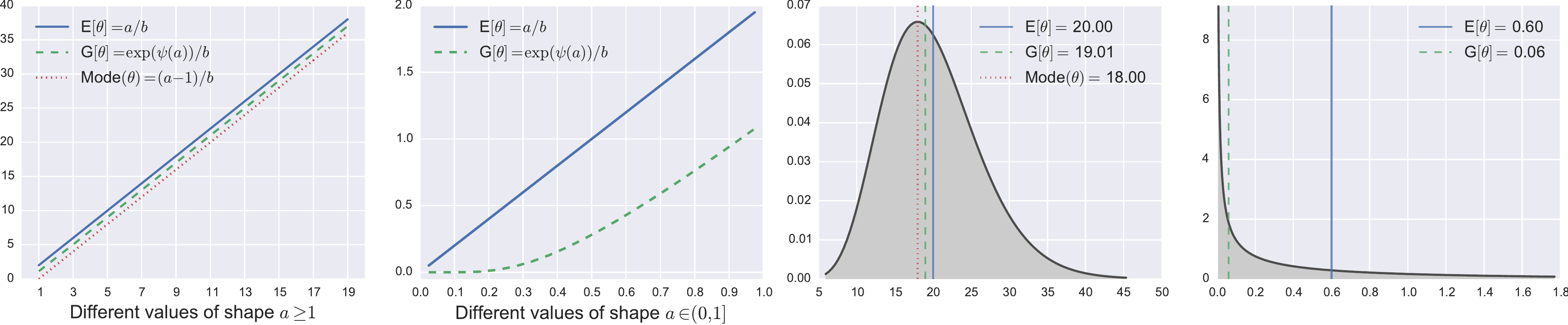}
  \end{center}
\vspace{-1em}
  \caption{ \scriptsize The mode, arithmetic expectation, and
    geometric expectation of a Gamma-distributed random variable
    $\theta$.  \emph{First}: The three quantities for different values
    of shape $a\geq 1$ (x axis) with rate $b=0.5$. All three grow
    linearly with $a$ and $\mathbbm{E}\left[\theta\right] \geq
    \mathbbm{G}\left[\theta\right] \geq \text{Mode}\,(\theta)$.
    \emph{Second}: Geometric and arithmetic expectations for different
    values of shape $a \in (0,1)$, where the mode is undefined, with
    rate $b=0.5$.  $\mathbbm{G}\left[\theta\right]$ grows more slowly
    than $\mathbbm{E}\left[\theta\right]$. This property is most
    apparent when $a < 0.4$.  \emph{Third}: pdf of a Gamma
    distribution with shape $a = 10$ and rate $b = 0.5$.  The three
    quantities are shown as vertical lines.  All three are close in
    the area of highest density, differing by about a half unit of
    inverse rate, i.e., $\frac{1}{2b} = 1$. \emph{Fourth}: pdf of a
    Gamma distribution with $a = 0.3$ and $b = 0.5$. The geometric and
    arithmetic expectations are shown as vertical lines (the mode is
    undefined). The two quantities differ greatly, with
    $\mathbbm{G}\left[\theta\right]$ much closer to zero and in an
    area of higher density. If these expectations were used as point
    estimates to predict the presence or absence of a rare
    event---e.g., $y = 0$ if $\hat{\theta} < 0.5$; otherwise $y =
    1$---they would yield different predictions.}
\vspace{-1em}
\end{figure*}

\section{Technical Discussion}
\label{sec:discussion}

Previous work on Bayesian Poisson matrix factorization
(e.g.,~\cite{cemgil09bayesian, paisley14bayesian, gopalan15scalable})
presented update equations for the variational parameters in terms of
auxiliary variables, known as \emph{latent sources}, and made no
explicit reference to geometric expectations. In contrast, we write
the update equations for Bayesian Poisson tensor factorization in the
form of equations~(\ref{eqn:gamma_update}) and
(\ref{eqn:delta_update}) in order to highlight their relationship to
Lee and Seung's multiplicative updates for non-negative tensor
factorization---a parallel also drawn by Cemgil in his paper
introducing Bayesian PMF~\cite{cemgil09bayesian}---and to show that
our update equations suggest a new way of making out-of-sample
predictions when using BPTF. In this section, we provide a discussion
of these connections and their implications.

When performing NTF by minimizing the generalized KL divergence of
reconstruction $\hat{\boldsymbol{Y}}$ from observed tensor
$\boldsymbol{Y}$ (which is equivalent to MLE for PTF), the
multiplicative update equation introduced by Lee and Seung for, e.g.,
$\theta^{(1)}_{ik}$ is
\begin{equation}
  \label{eq:lee_seung}
  \theta^{(1)}_{ik} := \theta^{(1)}_{ik} \frac{\sum_{j,a,t}
    \theta^{(2)}_{jk} \theta^{(3)}_{ak}
    \theta^{(4)}_{tk}\, \frac{y_{ijat}}{\hat{y}_{ijat}}
  }{\sum_{j,a,t} \theta^{(2)}_{jk} \theta^{(3)}_{ak} \theta^{(4)}_{tk}}.
  \end{equation}

These update equations sometimes converge to locally non-optimal
values when the observed tensor is very
sparse~\cite{gonzalez05accelerating,lin07convergence,chi12tensors}. This
problem occurs when factors are set to \emph{inadmissible zeros}; the
algorithm cannot recover from these values due to the multiplicative
nature of the update equations. Several solutions have been proposed
to correct this behavior when minimizing Euclidean distance. For
example, Gillis and Glineur~\cite{gillis08nonnegative} add a small
constant $\epsilon$ to each factor to prevent them from ever becoming
exactly zero. For KL divergence, Chi and Kolda~\cite{chi12tensors}
proposed an algorithm---Alternating Poisson Regression---that
``scooches'' factors away from zero more selectively (i.e., some
factors are still permitted to be zero).

In BPTF, point estimates of the latent factors are not estimated
directly. Instead, variational parameters for each factor, e.g.,
$\gamma^{(1)}_{ik}$ and $\delta^{(1)}_{ik}$ for factor
$\theta^{(1)}_{ik}$, are estimated. These parameters then define a
Gamma distribution over the factor as in
equation~(\ref{eq:q_for_single_theta}), thereby preserving uncertainty
about its value. In practice, this approach solves the instability
issues suffered by MLE methods, without any efficiency sacrifice. This
assertion is supported empirically by the out-of-sample predictive
performance results reported in section~\ref{sec:eval}, but can also
be verified by comparing the form of the update in
equation~(\ref{eq:lee_seung}) with those of the updates in
equations~(\ref{eqn:gamma_update}) and
(\ref{eqn:delta_update}). Specifically, if
equations~(\ref{eqn:gamma_update}) and~(\ref{eqn:delta_update}) are
substituted into the expression for the arithmetic expectation of a
single latent factor, e.g., $\mathbb{E}\left[ \theta^{(1)}_{ik}
  \right] = \frac{\gamma^{(1)}_{ik}}{\delta^{(1)}_{ik}}$, then the
resultant update equation is very similar to the update in
equation~(\ref{eq:lee_seung}):
\begin{equation*}
  \mathbb{E}_{Q}\left[ \theta_{ik}^{(1)}\right] := \frac{\alpha +
    \sum_{j,a,t} \mathbb{G}_Q\left[ \theta^{(1)}_{ik}
      \theta^{(2)}_{jk} \theta^{(3)}_{ak} \theta^{(4)}_{tk} \right]
    \frac{y_{ijat}}{\hat{y}_{ijat}}}{\alpha\,\beta^{(1)} +
    \sum_{j,a,t} \mathbb{E}_Q \left[ \theta_{jk}^{(2)} \theta_{ak}^{(3)}
      \theta_{tk}^{(4)} \right]},
  \end{equation*}
where $\hat{y}_{ijat} \equiv \sum_{k=1}^K \mathbb{G}_Q \left[
  \theta^{(1)}_{ik} \theta^{(2)}_{jk} \theta^{(3)}_{ak}
  \theta^{(4)}_{tk} \right]$. Pulling $\mathbb{G}_Q \left[
  \theta^{(1)}_{ik}\right]$ outside the sum in the numerator and
letting $\alpha \rightarrow 0$, yields
\begin{equation*}
  \mathbb{E}_{Q}\left[ \theta_{ik}^{(1)}\right] := \mathbb{G}_Q \left[ \theta_{ik}^{(1)}
    \right] \frac{
    \sum_{j,a,t} \mathbb{G}_Q\left[ 
      \theta^{(2)}_{jk} \theta^{(3)}_{ak} \theta^{(4)}_{tk} \right] \frac{y_{ijat}}{\hat{y}_{ijat}}
  }{
    \sum_{j,a,t} \mathbb{E}_Q \left[ \theta_{jk}^{(2)} \theta_{ak}^{(3)}
      \theta_{tk}^{(4)} \right]},
  \end{equation*}
which is exactly the form of equation~(\ref{eq:lee_seung}), except that
the point estimates of the factors are replaced with two kinds of
expectation. This equation makes it clear that the properties that
differentiate variational inference for BPTF from the Lee and Seung
updates for PTF are 1) the hyperparameters $\alpha$ and $\beta^{(1)}$
and 2) the use of arithmetic and geometric expectations of the factors
instead of direct point estimates.

Since the hyperparameters provide a form of implicit correction, BPTF
should not suffer from inadmissible zeros, unlike non-Bayesian PTF. It
is also interesting to explore the contribution of the geometric
expectations. The fact that each $\hat{y}_{ijat}$ is defined in terms
of a geometric expectation suggests that when constructing point
estimates of the latent factors from the variational distribution
(e.g., for use in prediction), the geometric expectation is more
appropriate than the arithmetic expectation (which is commonly used in
Bayesian Poisson matrix factorization) since the inference algorithm
is implicitly optimizing the reconstruction as defined in terms of
geometric expectations of the factors.

To explore the practical differences between geometric and arithmetic
expectations of the latent factors under the variational distribution,
it is illustrative to consider the form of $\textrm{Gamma}\,(\theta;
a, b)$. Most relevantly, the Gamma distribution is asymmetric, and its
mean (i.e., its arithmetic expectation) is greater than its mode. When
shape parameter $a \geq 1$, $\textrm{Mode}\,(\theta) =
\frac{(a-1)}{b}$; when $a < 1$, the mode is undefined, but most of the
distribution's probability mass is concentrated near zero---i.e., the
pdf increases monotonically as $\theta \rightarrow 0$. This property
is depicted in figure~2. In this scenario, the Gamma
distribution's heavy tail pulls the arithmetic mean away from zero and
into a region of lower probability.

The geometric expectation is upper-bounded by the arithmetic
expectation---i.e., $\mathbb{G}\left[ \theta \right] =
\frac{\exp{\left( \Psi(a) \right)}}{b} \leq \frac{a}{b} =
\mathbb{E}\left[ \theta \right]$. Unlike the mode, it is well-defined
for $a \in (0,1)$ and grows quadratically over this interval, since
$\exp{\left( \Psi(a) \right)} \approx \frac{a^2}{2}$ for $a \in
(0,1)$; in contrast, the arithmetic expectation grows linearly over
this interval. As a result, when $a < 1$, the geometric expectation
yields point estimates that are much closer to zero than those
obtained using the arithmetic expectation. When $a \geq 1$,
$\exp{\left( \Psi(a) \right)} \approx a - 0.5$ and the geometric
expectation is approximately equidistant between the arithmetic
expectation and the mode---i.e., $\frac{a}{b} \geq \frac{a - 0.5}{b}
\geq \frac{a-1}{b}$. These properties are depicted in
figure~7; the key point to take away from this figure is
that when $a < 1$, the geometric expectation has a much more probable
value than the arithmetic expectation, while when $a \geq 1$, the
geometric and arithmetic expectations are very close. This observation
suggests that the geometric expectation should yield similar or better
point estimates of the latent factors than those obtained using the
arithmetic expectation. In table~2, we provide a
comparison of the out-of-sample predictive performance for BPTF using
arithmetic and geometric expectations. Indeed, these results show that
the performance obtained using geometric expectations is either the
same as or better than the performance obtained instead using
arithmetic expectations.

\begin{table}[t] \label{fig:ari-vs-geo}
\scriptsize
\caption{\scriptsize Predictive performance obtained using geometric
  and arithmetic expectations. (The experimental design was identical
  to that used to obtain the results in table~1.)
  Using geometric expectations resulted in the same or better
  performance than that obtained using arithmetic expectations.}
\begin{center}
\begin{tabular}{l l l l l l l} \toprule
 & &  \multicolumn{2}{c}{BPTF-ARI} & \multicolumn{2}{c}{BPTF-GEO}\\ \cmidrule(r){3-4} \cmidrule(r){5-6}
 & Density  & MAE & HAM-Z & MAE & HAM-Z\\ \midrule
I-top-25 & 0.1217 &2.03 & 0.121 & {\bf 1.99} & {\bf 0.113}\\ %\midrule
G-top-25 & 0.2638 &8.96 & 0.3 & {\bf 8.94} & {\bf 0.292}\\ %\midrule
I-top-100 & 0.0264 &0.197 & 0.0236 & {\bf 0.178} & {\bf 0.0142}\\ %\midrule
G-top-100 & 0.0588 &1 & 0.0857 & {\bf 0.95} & {\bf 0.0682}\\ \midrule
I-top-25$^{c}$ & 0.0021 &0.0104 & 0.00163 & {0.0104} & {0.00161}\\ %\midrule
G-top-25$^{c}$ & 0.0060 &0.0414 & 0.00606 & {\bf 0.0412} & {\bf 0.00601}\\ %\midrule
I-top100$^{c}$ & 0.0004 &0.0011 & 5.03e-05 & {0.00109} & {4.97e-05}\\ %\midrule
G-top100$^{c}$ & 0.0015 &0.00804 & 0.000959 & {0.00803} & {0.000957}\\
\bottomrule
\end{tabular}
\end{center}
\end{table}

\section{Summary}

Over the past fifteen years, political scientists have engaged in an
ongoing debate about using dyadic events to study inherently
multilateral phenomena. This debate, as summarized by
Stewart~\cite{stewart14latent}, began with Green et al.'s
demonstration that many regression analyses based on dyadic events
were biased due to implausible independence
assumptions~\cite{green01dirty}. Researchers continue to expose such
biases, e.g., ~\cite{erikson14dyadic}, and some have even advocated
eschewing dyadic data on principle, calling instead for the
development of multilateral event data
sets~\cite{poast10misusing}. Taking the opposite viewpoint---i.e.,
that dyadic events can be used conduct meaningful analyses of
multilateral phenomena---other researchers, beginning with
Hoff~\cite{hoff04modeling}, have developed Bayesian latent factor
regression models that explicitly model unobserved dependencies as
occurring in some latent space, thereby controlling for their effects
in analyses. This line of research has seen an increase in interest
and activity over the past few
years~\cite{hoff13equivariant,stewart14latent,hoff14multilinear}.

In this paper, we too take this latter viewpoint, but instead of
focusing on latent factor models for regression, we present a Bayesian
latent factor model for predictive and exploratory data
analysis---specifically, for identifying and characterizing the
``complex dependence structures in international
relations''~\cite{king01proper} implicit in dyadic event data. Our
exploratory analysis revealed interpretable multilateral structures
that capture both persistent regional relations and temporally
localized anomalies. As evidenced empirically by our predictive
experiments and analytically by a comparison of our variational
inference algorithm with traditional algorithms for performing
non-negative tensor factorization, Bayesian Poisson tensor
factorization overcomes the instability issues exhibited by standard
non-negative tensor factorization methods when decomposing sparse,
dispersed count data. We provided additional analysis and empirical
results demonstrating that when constructing point estimates of the
latent factors from the variational distribution, the geometric
expectation is a more appropriate choice than the arithmetic
expectation. We therefore recommend its use in any subsequent work
involving variational inference for Bayesian Poisson matrix or tensor
factorization.

\section{Acknowledgments}

Thank you to Mingyuan Zhou, Brendan O'Connor, Brandon Stewart, Roy
Adams, David Belanger, Luke Vilnis, and Juston Moore for very helpful
discussions. This work was partially undertaken while Aaron Schein was
an intern at Microsoft Research New York City. This work was supported in part
by the UMass Amherst CIIR and in part by NSF grants \#IIS-1320219,
\#SBE-0965436, and \#IIS-1247664; ONR grant \#N00014-11-1-0651; and
DARPA grant \#FA8750-14-2-0009. Any opinions, findings, and
conclusions or recommendations expressed in this material are 
the
authors' and do not necessarily reflect those of the sponsor.


% \pagebreak

\bibliographystyle{abbrv}
\bibliography{ptf}

\begin{thebibliography}{10}

\bibitem{cemgil09bayesian}
A.~Cemgil.
\newblock Bayesian inference for nonnegative matrix factorisation models.
\newblock {\em Computational Intelligence and Neuroscience}, 2009.

\bibitem{chi12tensors}
E.~Chi and T.~Kolda.
\newblock On tensors, sparsity, and nonnegative factorizations.
\newblock {\em SIAM Journal on Matrix Analysis and Applications},
  33(4):1272--1299, 2012.

\bibitem{dorfman79formula}
R.~Dorfman.
\newblock A formula for the {G}ini coefficient.
\newblock {\em The Review of Economics and Statistics}, pages 146--149, 1979.

\bibitem{erikson14dyadic}
R.~Erikson, P.~Pinto, and K.~Rader.
\newblock Dyadic analysis in international relations: A cautionary tale.
\newblock {\em Political Analysis}, 22(4):457--463.

\bibitem{ermis14bayesian}
B.~Ermis and A.~Cemgil.
\newblock A {B}ayesian tensor factorization model via variational inference for
  link prediction.
\newblock arXiv:1409.8276, 2014.

\bibitem{gerner02conflict}
D.~Gerner, P.~Schrodt, R.~Abu-Jabr, and {\"{O}}.~Yilmaz.
\newblock Conflict and mediation event observations ({CAMEO}): A new event data
  framework for the analysis of foreign policy interactions.
\newblock Working paper.

\bibitem{gillis08nonnegative}
N.~Gillis and F.~Glineur.
\newblock Nonnegative factorization and the maximum edge biclique problem.
\newblock arXiv:0810.4225, 2008.

\bibitem{gonzalez05accelerating}
E.~Gonzalez and Y.~Zhang.
\newblock Accelerating the {L}ee-{S}eung algorithm for non-negative matrix
  factorization.
\newblock Technical Report TR-05-02, Department of Computational and Applied
  Mathematics, Rice University, 2005.

\bibitem{gopalan13efficient}
P.~Gopalan and D.~Blei.
\newblock Efficient discovery of overlapping communities in massive networks.
\newblock {\em Proceedings of the National Academy of Sciences}, 2013.

\bibitem{gopalan12scalable}
P.~Gopalan, S.~Gerrish, M.~Freedman, D.~Blei, and D.~Mimno.
\newblock Scalable inference of overlapping communities.
\newblock In {\em Advances in Neural Information Processing Systems
  Twenty-Five}, 2012.

\bibitem{gopalan15scalable}
P.~Gopalan, J.~Hofman, and D.~Blei.
\newblock Scalable recommendation with {P}oisson factorization.
\newblock In {\em Proceedings of the Thirty-First Conference on Uncertainty in
  Artificial Intelligence}, 2015.

\bibitem{green01dirty}
D.~Green, S.~Kim, and D.~Yoon.
\newblock Dirty pool.
\newblock {\em International Organization}, 55(2):441--468, 2001.

\bibitem{harshman70foundations}
R.~Harshman.
\newblock Foundations of the {PARAFAC} procedure: Models and conditions for an
  "explanatory" multimodal factor analysis.
\newblock {\em UCLA Working Papers in Phonetics}, 16:1--84, 1970.

\bibitem{hoff13equivariant}
P.~Hoff.
\newblock Equivariant and scale-free {T}ucker decomposition models.
\newblock arXiv:1312.6397, 2013.

\bibitem{hoff14multilinear}
P.~Hoff.
\newblock Multilinear tensor regression for longitudinal relational data.
\newblock arXiv:1412.0048, 2014.

\bibitem{hoff04modeling}
P.~Hoff and M.~Ward.
\newblock Modeling dependencies in international relations networks.
\newblock {\em Political Analysis}, 12(2):160--175, 2004.

\bibitem{king01proper}
G.~King.
\newblock Proper nouns and methodological propriety: Pooling dyads in
  international relations data.
\newblock {\em International Organization}, 55(2):497--507, 2001.

\bibitem{kolda08scalable}
T.~Kolda and J.~Sun.
\newblock Scalable tensor decompositions for multi-aspect data mining.
\newblock In {\em Proceedings of the Eighth IEEE International Conference on
  Data Mining}, pages 363--372, 2008.

\bibitem{lee99learning}
D.~Lee and S.~Seung.
\newblock Learning the parts of objects by non-negative matrix factorization.
\newblock {\em Nature}, 401:788--791, 1999.

\bibitem{leetaru13gdelt}
K.~Leetaru and P.~Schrodt.
\newblock {GDELT}: Global data on events, location, and tone, 1979--2012.
\newblock Working paper, 2013.

\bibitem{liang15codebook-based}
D.~Liang, J.~Paisley, and D.~Ellis.
\newblock Codebook-based scalable music tagging with {P}oisson matrix
  factorization.
\newblock In {\em Proceedings of the Fifteenth International Society for Music
  Information Retrieval Conference}, 2015.

\bibitem{lin07convergence}
C.~Lin.
\newblock On the convergence of multiplicative update algorithms for
  nonnegative matrix factorization.
\newblock {\em IEEE Transactions on Neural Networks}, 18(6):1589--1596, 2007.

\bibitem{marlin04collaborative}
B.~Marlin.
\newblock Collaborative filtering: A machine learning perspective.
\newblock Master's thesis, University of Toronto, 2004.

\bibitem{o'brien10crisis}
S.~O'Brien.
\newblock Crisis early warning and decision support: Contemporary approaches
  and thoughts on future research.
\newblock {\em International Studies Review}, 12(1):87--104, 2010.

\bibitem{paisley14bayesian}
J.~Paisley, D.~Blei, and M.~Jordan.
\newblock Bayesian nonnegative matrix factorization with stochastic variational
  inference.
\newblock In {\em Handbook of Mixed Membership Models and Their Applications}.
  Chapman and Hall/CRC, 2014.

\bibitem{poast10misusing}
P.~Poast.
\newblock ({M}is)using dyadic data to analyze multilateral events.
\newblock {\em Political Analysis}, 2010.

\bibitem{schein14inferring}
A.~Schein, J.~Paisley, D.~Blei, and H.~Wallach.
\newblock Inferring polyadic events with {P}oisson tensor factorization.
\newblock In {\em Proceedings of the NIPS 2014 Workshop on "Networks: From
  Graphs to Rich Data"}, 2014.

\bibitem{singer94correlates}
D.~Singer and M.~S. (producers).
\newblock Correlates of war project: International and civil war data,
  1816--1992 (computer file).
\newblock Inter-University Consortium for Political and Social Research
  (distributor), 1994.

\bibitem{stewart14latent}
B.~Stewart.
\newblock Latent factor regressions for the social sciences.
\newblock Working paper, 2014.

\bibitem{tucker66some}
L.~Tucker.
\newblock Some mathematical notes on three-mode factor analysis.
\newblock {\em Psychometrika}, 31(3):279--311, 1966.

\bibitem{ward13comparing}
M.~Ward, A.~Beger, J.~Cutler, M.~Dickenson, C.~Doorff, and B.~Radford.
\newblock Comparing {GDELT} and {ICEWS} event data.
\newblock Working paper, 2013.

\bibitem{welling01positive}
M.~Welling and M.~Weber.
\newblock Positive tensor factorization.
\newblock {\em Pattern Recognition Letters}, 22(12):1255--1261, 2001.

\bibitem{wikipedia15cruise}
Wikipedia.
\newblock Cruise missile strikes on {A}fghanistan and {S}udan ({A}ugust 1998).
\newblock Accessed June 8, 2015.

\bibitem{wikipedia14embassy}
Wikipedia.
\newblock Embassy of {E}cuador, {L}ondon.
\newblock Accessed October 30, 2014.

\bibitem{wikipedia15japanese}
Wikipedia.
\newblock {J}apanese {I}raq reconstruction and support group.
\newblock Accessed June 8, 2015.

\bibitem{wikipedia15jyllands-posten}
Wikipedia.
\newblock {J}yllands-{P}osten {M}uhammad cartoons controversy.
\newblock Accessed June 8, 2015.

\bibitem{wikipedia15six-party}
Wikipedia.
\newblock Six-party talks.
\newblock Accessed June 8, 2015.

\bibitem{zhou15negative}
M.~Zhou and L.~Carin.
\newblock Negative binomial process count and mixture modeling.
\newblock {\em IEEE Transactions on Pattern Analysis and Machine Intelligence},
  37(2):307--320, 2015.

\bibitem{zhou11beta-negative}
M.~Zhou, L.~Hannah, D.~Dunson, and L.~Carin.
\newblock Beta-negative binomial process and {P}oisson factor analysis.
\newblock arXiv:1112.3605, 2011.

\end{thebibliography}

% ACM needs 'a single self-contained file'!

% That's all folks!
\end{document}